\documentclass[twoside,11pt]{article}

\usepackage{verbatim, amsmath, graphicx, mathtools}
\usepackage{array}
\usepackage{amsthm}
\usepackage{graphicx}
\usepackage[ruled,vlined,linesnumbered,resetcount]{algorithm2e}
\usepackage{xcolor} 

\graphicspath{{./}}

\DeclareMathOperator*{\argmax}{argmax} 

\mathchardef\mhyphen="2D

\usepackage{jmlr2e}

\newcommand{\ubar}[1]{\text{\b{$#1$}}}

\jmlrheading{21}{2020}{1-?}{01/20}{?/19}{garton2019}{Nathaniel Garton and Jarad Niemi and Alicia Carriquiry}

\ShortHeadings{Knot Selection in Sparse GPs}{Garton, Niemi, and Carriquiry}
\firstpageno{1}

\begin{document}

\title{Knot Selection in Sparse Gaussian Processes}

\author{\name Nathaniel Garton \email nmgarton@iastate.edu \\
       \addr Department of Statistics\\
       Iowa State University\\
       Ames, IA 50011, USA
       \AND
       \name Jarad Niemi \email niemi@iastate.edu \\
       \addr Department of Statistics\\
       Iowa State University\\
       Ames, IA 50011, USA
       \AND
       \name Alicia Carriquiry \email alicia@iastate.edu \\
       \addr Department of Statistics\\
       Iowa State University\\
       Ames, IA 50011, USA}

\editor{}

\maketitle

\begin{abstract}
Knot-based, sparse Gaussian processes have enjoyed considerable success as scalable 
approximations to full Gaussian processes. Problems can occur, however, when knot selection
is done by optimizing the marginal likelihood. For example, the marginal likelihood surface is 
highly multimodal, which can cause suboptimal knot placement where some knots serve 
practically no function. This is especially a problem when many more knots are used
than are necessary, resulting in extra computational cost for little to no gains in accuracy.
 We propose a one-at-a-time knot selection algorithm to select both the number and placement of knots.
Our algorithm uses Bayesian optimization to efficiently propose knots that are 
likely to be good and largely avoids the pathologies encountered when using the marginal likelihood
as the objective function. We provide empirical results showing improved accuracy and 
speed over the current standard approaches. 
\end{abstract}


\section{Introduction}\label{s:intro}


Gaussian processes (GPs) are flexible tools for doing nonparametric Bayesian inference of latent functions. However, a well known limitation of such models is that computations involve the inversion and determinant of an $N \times N$ covariance matrix where $N$ is the number of data points. Thus, computation time scales as $\mathcal{O}(N^3)$, and storage of the $N\times N$ covariance matrix demands memory $\mathcal{O}(N^2)$. 

Several approximations to full Gaussian processes have been proposed for which computation time and memory scale linearly in the size of the data set \citep{smola2001, williams2001, seeger2003, snelson2006, banerjee2008, finley2009, datta2016}, many of which are described in Chapter 8 of \cite{rasmussen2006} and \cite{candela2005}. 
All of these proposed methods result in a GP with a sparse inverse covariance matrix, 
also called the precision matrix,
for which many take advantage of the well-known Sherman-Woodbury-Morrison 
matrix inversion formula 
\citep{smola2001, williams2001, snelson2006, banerjee2008, finley2009}. 

One popular method has been independently introduced in \cite{snelson2006} and in the spatial statistics literature in \citep{banerjee2008, finley2009} which is commonly called either the Fully Independent Conditional (FIC) model \citep{candela2005} or a (modified) predictive process model \citep{banerjee2008, finley2009}. 
This approximation involves selecting a set of $K$ locations, 
$x^{\dagger} = \left\{x^{\dagger}_1,\ldots,x^{\dagger}_K\right\}$ where $K\ll N$, 
and assumes that the data at the observed locations $x = \left\{x_1,\ldots,x_N\right\}$ are 
conditionally independent given the latent function values at $x^{\dagger}$.
These locations, $x^{\dagger}$, are often referred to as knots \citep{banerjee2008, finley2009}, pseudo-inputs \citep{snelson2006}, 
or inducing points/inputs \citep{candela2005}. Such locations can be treated as a subset of the observed data locations \citep{seeger2003,titsias2009, cao2013} but they need not be \citep{snelson2006, finley2009}.

In such a model, the question of how to select knots becomes of interest, and it has been noticed that evenly spaced knots often do not result in optimal performance \citep{finley2009}. 
\cite{snelson2006} demonstrated that, when the likelihood is Gaussian, 
one can treat the knots as parameters and choose them to optimize the 
likelihood using gradient ascent. 
Others have used variational inference (VI) with specific approximate posteriors that give rise to similar knot-based sparse models for Gaussian process regression and classification \citep{titsias2009, hensman2015}. In the VI framework, knots are optimized as variational parameters in the approximating posterior. That is, they are chosen to minimize the Kullback-Leibler divergence between the approximate, sparse posterior and the full GP posterior. Others have proposed a variety of ways to select knots including greedy approaches with alternative objective functions, for example predictive variance \citep{finley2009} or ``information gain" \citep{seeger2003}. When the likelihood is non-Gaussian, exact inference becomes intractable and two options for optimizing knots are to optimize an analytical approximation to the marginal likelihood with respect to the knots or use VI \citep{hensman2015, lobato2016}. 

To our knowledge, no methods have been proposed and tested which automatically select the number of knots. \cite{titsias2009} suggested the possibility of alternating between greedily adding knots and optimizing covariance parameters, but no experimental work implementing this suggestion appears to have been published. 
A naive approach to selecting the number of knots is to include as many as is 
computationally feasible. 
As we will show, such an approach may produce needlessly complex solutions if there are sparser 
approximations with competitive accuracy. 
Perhaps more importantly, while the number of knots impacts the quality of the approximation, we observe that when the number of knots is large or knots are initialized poorly, simultaneous knot optimization can converge to suboptimal solutions. This performance is explained in \cite{bauer2016} and manifests itself as ``clumping" where several knots are in close proximity
in the likelihood optimization.
We provide an example similar to ones found in \cite{bauer2016} of how continuous optimization of even one knot can be difficult. 
We argue that an intuitive reason for these phenomena has to do with how an individual knot
affects the implied marginal covariance structure at the observed data 
locations. 


To address these issues, we propose a method for selecting the number and location of knots in GP models where the objective function is the marginal likelihood. When the data are non-Gaussian, we propose to use the Laplace approximation to the marginal likelihood.
Like much of the recent work involving FIC approximations, 
we assume a covariance function that is differentiable in the knots \citep{snelson2006, hensman2015, lobato2016}. 
Our algorithm works by sequentially adding and optimizing single knots 
alongside covariance parameters. We demonstrate superior accuracy of this approach relative to the simultaneous optimization of all knots and covariance parameters, which largely results from
the mitigation of optimization issues apparent on several benchmark data sets. 
Further, while computation time for both our algorithm and simultaneous knot optimization scales linearly 
with the number of data points, 
our approach winds up in practice being faster than simultaneous optimization. 
The reason for this is that the number of derivatives required for doing 
gradient ascent is independent of $K$, 
making gradient ascent steps $\mathcal{O}(K)$ cheaper. 
Our approach does require some extra computational effort in proposing knot 
locations, but our experiments show that this extra cost is outweighed by the 
savings related to cheaper gradient evaluations.


Our proposal for doing knot selection with non-Gaussian data makes it applicable for many different types of data. Any type of data that can be modeled with a generalized linear mixed model (GLMM), for example Bernoulli, Poisson, gamma, is amenable to the methods we propose here. Further, software exists to use Laplace approximations for inference in latent GP models where the data are Student's t-distributed, Weibull, or log-Gaussian \citep{rasmussen2010,vanhatalo2013}, to name a few.


The rest of the paper is organized as follows. Section \ref{s:lgpm} introduces models incorporating latent Gaussian processes. In Section \ref{s:fic} we introduce the sparse, knot-based model that we will consider for the remainder of the paper. Section \ref{s:selectk} provides a toy example to motivate the need for selecting the number of knots and for care in their placement. Next, in Section \ref{s:oat} we propose our one-at-a-time knot selection algorithm. In Section \ref{s:experiments} we compare our algorithm to simultaneous optimization of a preselected number of knots as well as to full GPs on three benchmark data sets and show superior performance. In Section \ref{s:discussion} we conclude by discussing possible improvements to our algorithm and its use with other types of inference algorithms. 


\section{Latent Gaussian Processes}\label{s:lgpm}
We assume that we have $N$ observations, $(y_i,x_i^{\top})$, 
from a data set where each $y_i$ is the target of interest, 
and the values $x_i$ are vectors of input variables where 
$x_i \in \mathcal{X} \subset \mathbb{R}^d$. 
We suppose that over $\mathcal{X}$ there is an unobservable, 
real-valued function $f:\mathcal{X}\to \mathbb{R}$ 
taking values $f(x_i)$. 
We further suppose that the values of this function control the mean of the (conditional) distribution of the target random variable $Y_i$. That is, for some chosen link function $g(\cdot)$, 
we have $E\left[ Y_i|f(x_i) \right] = g(f(x_i))$. 
We can use a GP to put a prior distribution on the latent function, 
which we denote as $f(x) \sim \mathcal{GP}(m(x), k_{\theta}(x,x'))$.

A Gaussian process is specified by a mean function $m(x)$ and 
a covariance function $k_{\theta}(x,x')$, 
which we assume is parameterized by $\theta$. 
A GP, by definition, is a collection of random variables such that any finite 
subcollection $f_x = (f(x_1), ...,f(x_M))^{\top} \sim \mathcal{N}_M(m_x, \Sigma_{xx})$ 
where 
$m_x = (m(x_1), ..., m(x_M))^{\top}$ and the $ij$-th element of 
$\Sigma_{xx}(i,j) = k_{\theta}(x_i, x_j)$.

The models that we consider require a joint distribution that can be written as the product 
$p(y|f_x)p(f_x|\theta)$. 
Here, $p(y|f_x)$ is the distribution for the data, $y = (y_1, ..., y_N)^{\top}$, 
conditional on the values of an unobserved Gaussian process 
having distribution $p(f_x|\theta)$. 
We assume that each random variable $Y_i$ is independent conditional on $f(x_i)$, 
the function value at $x_i$.


\section{FIC Approximation}\label{s:fic}
The FIC model assumes that, 
conditional on the values of a GP at $K$ knots, 
$x^{\dagger} = \left\{x_1^\dagger, \ldots, x_K^\dagger \right\}$, 
all other latent function values are independent. 
In defining the FIC approximation mathematically, 
it will be helpful to first define the joint distribution of the unobserved 
function at the observed input locations, $f_x$, 
as well as the values at the potentially unobserved knot locations 
which we will call $f_{x^{\dagger}}$. 
This joint distribution can be written as 
$p(f_x,f_{x^{\dagger}}|\theta) = p(f_x|f_{x^{\dagger}}, \theta)p(f_{x^{\dagger}}|\theta)$. 
The FIC approximation considers an approximation to the full conditional distribution $p(f_x| f_{x^{\dagger}}, \theta)$ where


\begin{equation}
\begin{array}{lll}
f_x| f_{x^{\dagger}} , \theta & \sim & \mathcal{N}\left( m_x + \Sigma_{xx^{\dagger}} \Sigma_{x^{\dagger}x^{\dagger}}^{-1}(f_{x^{\dagger}} - m_{x^{\dagger}}) , \Lambda \right) \\
f_{x^{\dagger}}| \theta & \sim & \mathcal{N}(m_{x^{\dagger}}, \Sigma_{x^{\dagger}x^{\dagger}}).
\end{array}
\end{equation}

\noindent The marginal distribution of $f_x$ 
is then 
\begin{equation}
f_x|x^{\dagger}, \theta \sim \mathcal{N}(m_{x}, \Lambda + \Sigma_{x x^{\dagger}} \Sigma_{x^{\dagger}x^{\dagger}}^{-1} \Sigma_{x^{\dagger}x}),
\end{equation}
where $\Lambda = \text{diag}(\Sigma_{xx} - \Sigma_{x x^{\dagger}} \Sigma_{x^{\dagger}x^{\dagger}}^{-1} \Sigma_{x^{\dagger}x})$. 
Thus, the FIC approximation results in a marginal distribution for $f_x$ where 
the marginal means and variances are the same as they would have been if we were 
to use the full GP model, 
but covariances are exclusively controlled by the knots.

\subsection{Gaussian data}\label{s:gaussian_example}

In the case of Gaussian data, 
we assume an identity link function, $g(x) = x$
and conditionally Gaussian response variables, where
$Y|f_x \sim \mathcal{N}(f_x, \tau^2 I)$, where $Y = (Y_1, ..., Y_N)^{\top}$
is the vector of response random variables at the observed input locations.
For notational compactness, we will define $\Psi_{xx} \equiv \Lambda + \Sigma_{x x^{\dagger}} \Sigma_{x^{\dagger}x^{\dagger}}^{-1}\Sigma_{x^{\dagger}x}$. The marginal distribution for $Y|x, x^{\dagger}, \theta$ is then 

\begin{equation}
Y|x, x^{\dagger}, \theta \sim \mathcal{N}(m_{x}, \tau^2I + \Psi_{xx}).
\end{equation}

\noindent This marginal distribution can be optimized with respect to the knots $x^{\dagger}$ as well as covariance parameters $\theta$.

The posterior distribution for an unobserved response vector, $\tilde{Y}$, at inputs $\tilde{x} = \{ \tilde{x}_1, ...,  \tilde{x}_M \}$ is Gaussian with conditional expectation 

\begin{equation*}
E\left[ \left.\tilde{Y}\right|Y \right] = 
m_{\tilde{x}} +
\Sigma_{\tilde{x}x^{\dagger}} \Sigma_{x^{\dagger}x^{\dagger}}^{-1} \left( E\left[ f_{x^{\dagger}} | Y \right] - m_{x^{\dagger}} \right)
\end{equation*}

\noindent and conditional variance

\begin{equation*}
V\left[\left.\tilde{Y}\right|Y\right] = (\tau^2I + \Sigma_{\tilde{x}\tilde{x}}) 
- \Sigma_{\tilde{x}x^{\dagger}} \left( \Sigma_{x^{\dagger}x^{\dagger}}^{-1} - \Sigma_{x^{\dagger}x^{\dagger}}^{-1} V\left[f_{x^{\dagger}}|Y\right] \Sigma_{x^{\dagger}x^{\dagger}}^{-1}  \right) \Sigma_{x^{\dagger}\tilde{x}},
\end{equation*}

\noindent where 

\begin{equation*}
f_{x^{\dagger}}|Y \sim \mathcal{N}(m_{x^{\dagger}} + \Sigma_{x^{\dagger}x} (\Psi_{xx} + \tau^2I)^{-1}(y - m_x)  ,  \Sigma_{x^{\dagger}x^{\dagger}} - \Sigma_{x^{\dagger}x} (\Psi_{xx} + \tau^2I)^{-1} \Sigma_{xx^{\dagger}}).
\end{equation*}

\noindent Note that the covariance between any random variables $Y_i$ and $f_{x^{\dagger}_j}$ is just $k_{\theta}(x_i,x^{\dagger}_j)$. Computation of $(\tau^2I+\Psi_{xx})^{-1}$ can be done through the Sherman-Woodbury-Morrison matrix inversion formula.

\subsection{Non-Gaussian Data \& the Laplace Approximation}\label{s:nongaussian_data}
In the non-Gaussian data case, the marginal likelihood of 
$Y|x, x^{\dagger},\theta$ cannot be computed analytically. 
Hence we utilize the Laplace approximation to the integral 
$\int{p(y|f_x)p(f_x|x^{\dagger}, \theta) df_x}$. 
Our main reasoning for choosing this over other methods is the Laplace 
approximation's relative superiority as an estimate of the true marginal 
likelihood for the purpose of hyperparameter selection \citep{nickish2008}. 
Following notation in \cite{rasmussen2006}, we define the function 
$\psi(f_x) \equiv \log p(y|f_x) + \log p(f_x|x^{\dagger}, \theta)$. 
We proceed by taking the second order Taylor expansion of $\psi(f_x)$ around its
mode, say $\hat{f}_x$. This results in the following expression

\begin{equation}
\psi(f_x) \approx \psi\left(\hat{f}_x\right) + 
\frac{1}{2}\left(f_x - \hat{f}_x\right)^{\top}\left[ \nabla^2 \psi(f_x)|_{\hat{f}_x} \right]\left(f_x - \hat{f}_x\right),
\end{equation}
where $\nabla^2 \psi(f_x)|_{\hat{f}_x}$ is the Hessian of the function $\psi(f_x)$ evaluated at $f_x = \hat{f}_x$. If $p(y|f_x)$ is log-concave in $f_x$ 
(and it is for the distributions and link functions we consider), 
then $\psi(f_x)$ is unimodal \citep{rasmussen2006}. The Laplace approximation then results in an approximation to the marginal likelihood given by 

\begin{align*}
\int{p(y|f_x)p(f_x|x^{\dagger}, \theta) df} &\approx \int{ \text{exp}\left( \psi\left(\hat{f}_x\right) + \frac{1}{2}\left(f_x - \hat{f}_x\right)^{\top}\left[ \nabla^2 \psi(f_x)|_{\hat{f}_x} \right]\left(f_x - \hat{f}_x\right) \right) df_x } \\
&= (2\pi)^{n/2} e^{\psi\left(\hat{f}_x\right)} \Big\lvert-\nabla^2 \psi(f_x)|_{\hat{f}_x}\Big\rvert^{-1/2}.
\end{align*}
We denote this approximation as $q(y|x, x^{\dagger}, \theta)$. This integral approximation also corresponds to an approximation to the posterior distribution given by
\[ p(f_x|y,x^{\dagger}, \theta) \approx \mathcal{N}\left(\hat{f}_x, \left[-\nabla^2 \psi(f_x)|_{\hat{f}_x} \right]^{-1}\right).\] 
Comprehensive details of the Laplace approximation for a full GP model with a Bernoulli likelihood,
including relevant derivatives, a Newton-Raphson algorithm for finding the mode of 
$\psi(f_x)$, and derivatives of $q(y|x, x^{\dagger}, \theta)$ with respect to kernel 
hyperparameters can be found in \cite{rasmussen2006}. Our implementations
for full GP models follow those in \cite{rasmussen2006} exactly. Additional implementational 
details of the Laplace approximation in the FIC model for Poisson data, such as 
the necessary modifications to derivative calculations can be 
found in \cite{vanhatalo2010}, and our implementation is very similar. These 
details are easily modified for other data distributions. 

To calculate point and interval estimates for the latent function under FIC models we use formulas which are provided below. 
These formulas are similar to the Gaussian case, 
but they now involve the Hessian and mode of the function $\psi(f_x)$. 
Recall that the Gaussian approximation to $p(f_x|Y, x^{\dagger}, \theta)$ is given by

\begin{equation*}
\mathcal{N}\left(\hat{f}_x, \left[-\nabla^2 \psi(f_x)|_{\hat{f}_x} \right]^{-1} \right).
\end{equation*}

\noindent The posterior predictive distribution of an unobserved latent function value $f_{\tilde{x}}$ at unobserved input location $\tilde{x}$ is Gaussian with expectation 

\begin{equation*}
E\left[ f_{\tilde{x}}|Y \right] = \Sigma_{\tilde{x}x^{\dagger}} \Sigma_{x^{\dagger}x^{\dagger}}^{-1} \left( E\left[ f_{x^{\dagger}} | Y \right] - m_{x^{\dagger}} \right) + m_{\tilde{x}}
\end{equation*}

\noindent and conditional variance

\begin{equation*}
V\left[f_{\tilde{x}}|Y\right] = \Sigma_{\tilde{x} \tilde{x}} - \Sigma_{\tilde{x}x^{\dagger}} \left( \Sigma_{x^{\dagger}x^{\dagger}}^{-1} - \Sigma_{x^{\dagger}x^{\dagger}}^{-1} V\left[f_{x^{\dagger}}|Y\right] \Sigma_{x^{\dagger}x^{\dagger}}^{-1}  \right) \Sigma_{x^{\dagger}\tilde{x}},
\end{equation*}

\noindent where $f_{x^{\dagger}}|Y$ has the following distribution,

\begin{equation*}
\begin{multlined}[b] f_{x^{\dagger}}|Y \sim \mathcal{N}\bigg(\Sigma_{x^{\dagger}x} \Psi_{xx}^{-1}(\hat{f}_x - m_{x}) + m_{x^{\dagger}}, \\ \Sigma_{x^{\dagger}x^{\dagger}} - \Sigma_{x^{\dagger}x} \left[ \Psi_{xx}^{-1} - \Psi_{xx}^{-1} \left(-\nabla^2 \psi(f_x)|_{\hat{f}_x} \right)^{-1} \Psi_{xx}^{-1} \right] \Sigma_{x x^{\dagger}} \bigg) \end{multlined}.
\end{equation*}

\noindent 
The major difference between this case and the Gaussian case comes from the fact that the (approximate) distribution of $f_{x^{\dagger}}|Y$ isn't readily available from the initial model specification. 
Thus, we first recognize that, due to our Gaussian posterior approximation, the form of $f_{x^{\dagger}}|Y$ is Gaussian. 
To see this, note that the exact posterior  factors as 
$p(f_{x^{\dagger}},f_x|Y) = p(f_{x^{\dagger}}|f_x)p(f_x|Y)$. 
Due to the fact that we are using a Gaussian approximation of $p(f_x|Y)$ as well the
form of the dependency of $p(f_{x^{\dagger}}|f_x)$ on $f_x$, our 
approximation of $p(f_{x^{\dagger}}, f_x|Y)$ is Gaussian. By the 
marginalization property of the Gaussian distribution, our approximation to 
$p(f_{x^{\dagger}}|Y)$ is Gaussian. Full specification of this distribution 
requires computing $E\left[ f_{x^{\dagger}}|Y \right]$ and 
$V\left[f_{x^{\dagger}}|Y\right]$. Doing this requires using laws of iterated 
expectation and variance where 

\begin{equation*}
E\left[ f_{x^{\dagger}}|Y \right] = E\left[ E\left[ f_{x^{\dagger}}|f_x \right] |Y \right],
\end{equation*}

\noindent and

\begin{equation*}
V\left[ f_{x^{\dagger}}|Y \right] = E\left[ V\left[ f_{x^{\dagger}}|f_x \right] |Y \right] + V\left[E\left[ f_{x^{\dagger}}|f_x \right] |Y\right],
\end{equation*}

\noindent which is now straightforward to compute.

\section{Importance of Selecting $K$}\label{s:selectk}

The motivation for selecting $K$ using a data driven approach is two-fold. First, there are situations in practice where $K$ needn't be as large as the computational budget allows to produce equally accurate predictions. For example, the top left and right panels of Figure 1 in \cite{titsias2009} 
show point and interval predictions from sparse GPs with 15 knots that closely resemble those of the full GP, 
especially within the range of the observed inputs. 

\begin{figure}[htbp!]
\centering
\includegraphics[width = 0.6\textwidth, height = 0.25\textheight]{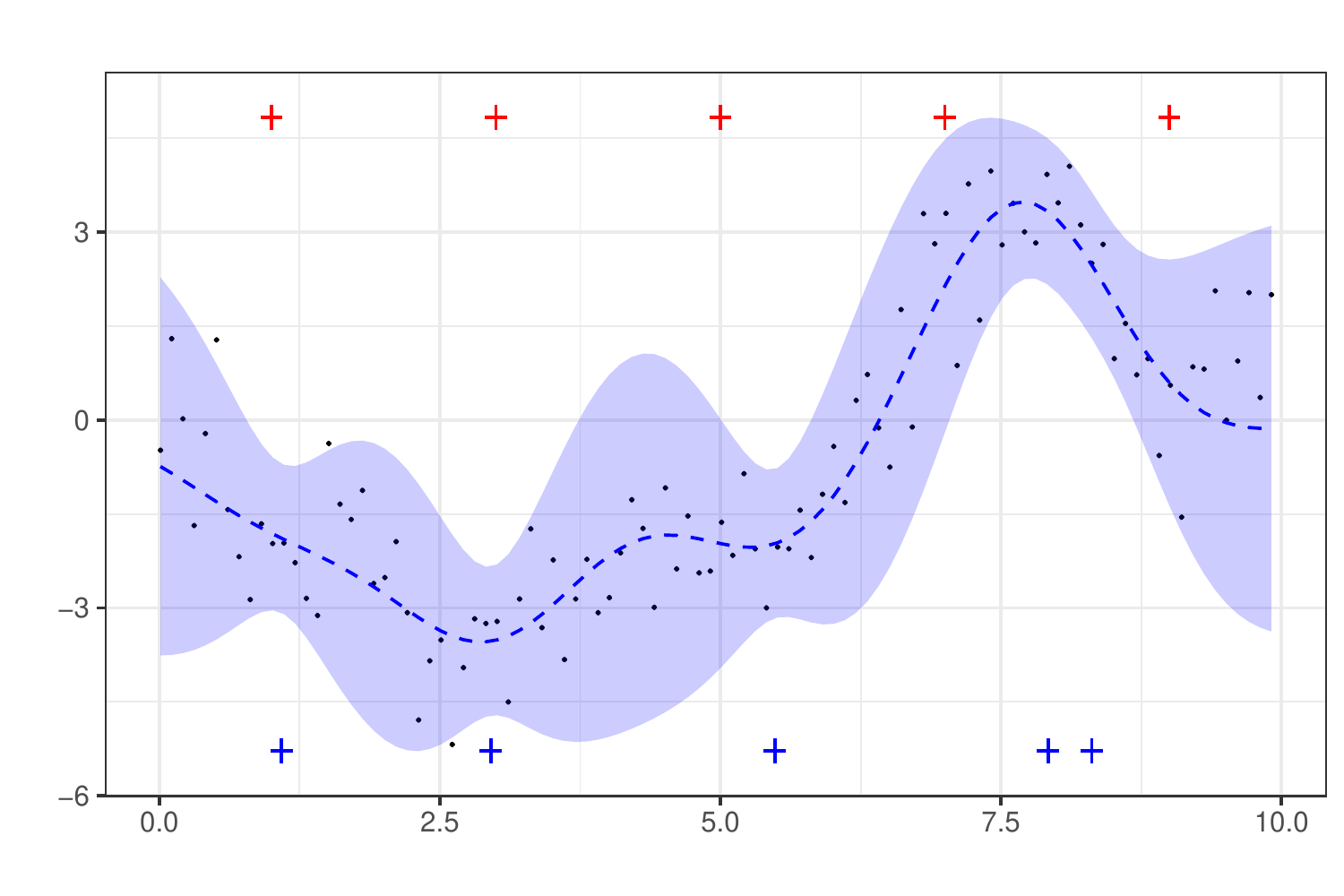}
\caption{Conditionally Gaussian data (black dots) with initial and estimated knot 
locations (red and blue $+$, respectively), 
estimated mean (dashed blue line), and 
pointwise 95\% credible intervals (shaded region). 
}
\label{fig:normal_example}
\end{figure}

Secondly, as noted in \cite{bauer2016}, optimizing knots using the (potentially approximate) marginal likelihood as the objective function is generally packed with suboptimal local maxima. 
To see this, Figure \ref{fig:normal_example} provides an example based on simulated Gaussian data and shows the selected knot locations, 
predictive mean and pointwise 95\% credible intervals from a fitted FIC model with five knots.

\begin{figure}[htbp!]
\centering
 \includegraphics[width = 0.6\textwidth, height = 0.25\textheight]{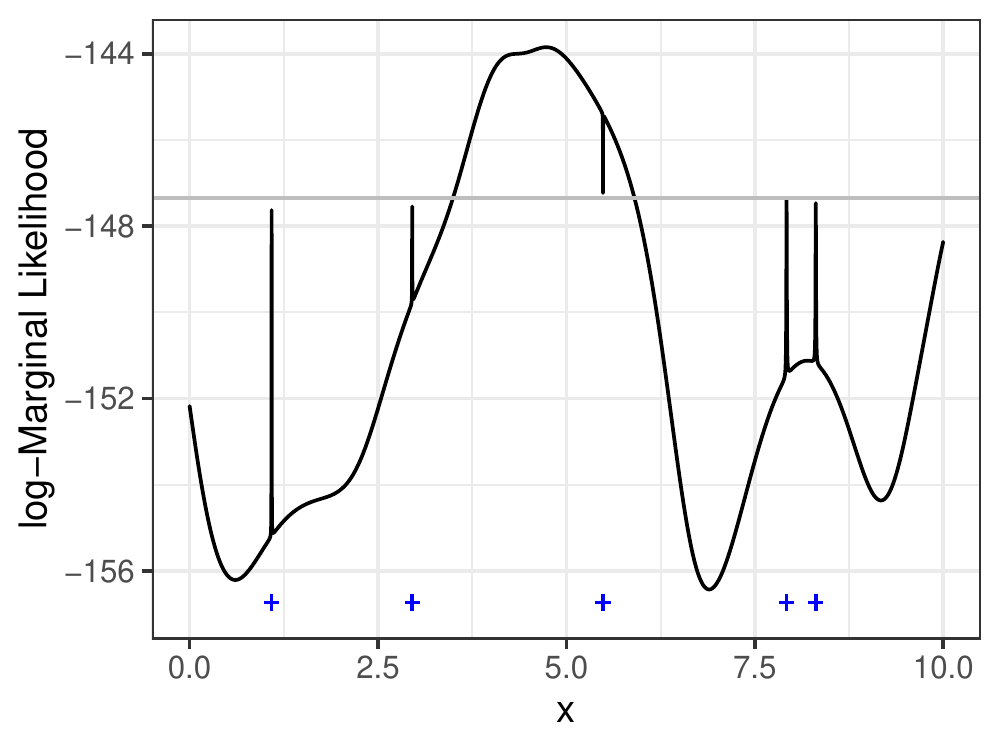}
 \caption{Log-marginal likelihood for a single, sixth knot with first five
 knots (blue $+$) and log-marginal likelihood for the model with the first
 five knots (horizontal gray line).
 }
 \label{fig:norm_marginal_ll}
\end{figure}

Additionally, the marginal likelihood for this model and these data as a 
function of a sixth knot is provided in Figure \ref{fig:norm_marginal_ll}. 
This figure shows that as the sixth knot approaches one of the five other knots 
in the model, the marginal likelihood gets arbitrarily close to the value 
it would take if the model did not include that knot at all. \cite{bauer2016}
provide a proof in their supplementary information that this is the case. 
This, ultimately, is unsurprising because the only role of the knots is to 
control the implied marginal covariances of the Gaussian prior on the latent function at 
the observed data locations. 
When two knots are arbitrarily close to each other, 
the covariances between the latent function at observed data locations are 
effectively no different than they would be if only one knot were at that location.  

The unfortunate effect of this, however, 
is that there are local optima near every one of the five knots:
either a local maxima or a local minima. \cite{bauer2016} note that in the absence
of a small noise variance or ``jitter" added to the latent function, these optima 
are only a problem theoretically. However, with the small noise variance, optimization
challenges arise.
Consider, for example, the middle knot near 5 on the x-axis in Figure \ref{fig:norm_marginal_ll}. There is a local minimum at this location which, because of the shape of the marginal likelihood to the right of this knot, forces a suboptimal local maximum immediately to the right. Alternatively, consider the farthest knot to the right. Here, there is a suboptimal local maximum because in the near vicinity of this knot the marginal likelihood is actually less than it is in the five knot model. In addition to this, there are large regions of the x-axis which, if chosen to be the starting value for the sixth knot during a gradient ascent optimization, would result in the knot getting stuck at the same location as one of the other five. Lastly, these local optima can cause serious numerical issues when trying to invert the matrix $\Sigma_{x^{\dagger}x^{\dagger}}$, and due to the extreme curvature of the marginal likelihood near the five knots, gradient based optimization methods may oscillate for a large number of iterations without every reaching a reasonable convergence tolerance. Arguably these problems get worse as we try to include more knots as there are simply more suboptimal local maxima. 

\section{One-at-a-Time (OAT) Knot Selection}\label{s:oat}
The above problems motivate us to explore other strategies for optimizing knots using the marginal likelihood. We propose a sequential knot selection scheme in which the marginal likelihood is never optimized with respect to more than one knot. We do this so that we can focus on intelligently proposing new knots that avoid bad local maxima. At a high level, we propose to first optimize the marginal likelihood with respect to covariance parameters only for a small initial selection of knots. We consider this an initialization step. 
Next, we propose a new knot which results in the maximum value of the marginal likelihood over a set of carefully selected candidate locations.
We denote the proposal function as $J(\cdot)$.
Finally, both the new knot and covariance parameters are jointly optimized using gradient ascent. We repeat these steps until either we have met our computational budget in terms of the number of knots or the improvement to the marginal likelihood falls below a pre-determined threshold. 

The OAT knot selection algorithm is shown schematically as Algorithm \ref{a:oat}.
This algorithm shares similarities with several other knot selection algorithms. 
Much like these algorithms, ours is greedy, selecting each knot in sequence to optimize some objective function \citep{seeger2003,finley2009, cao2013}. 
However, most algorithms alternate between selection of knots and optimization of kernel hyperparameters. This is the case in \cite{seeger2003} and \cite{finley2009}, but in those algorithms there is not a well-defined convergence criteria as the objective functions for knot selection and hyperparameter optimization are different. \cite{cao2013} use only one objective function, but the number of knots is predefined. 
Instead, our algorithm finds a sequence of best possible models given 
that the knots in the previously selected best model are fixed. 
\begin{algorithm}[H]
\SetAlgoLined
\textbf{Initialize:} $x^{\dagger} = \{x^{\dagger}_i \}_{i = 1}^{K_{I}}$ \;
$\hat{\theta} = \argmax_{\theta} p(y|x, x^{\dagger}, \theta)$ \;
\Repeat{$|x^{\dagger}| = K_{max}$ or convergence}{
propose new knot $x^{\dagger^*} \leftarrow J(y , x, x^{\dagger}, \hat{\theta})$ \;
$(\hat{x}^{\dagger^*}, \hat{\theta}) = \argmax_{(x^{\dagger^*}, \theta)} p(y|x, \{x^{\dagger}, x^{\dagger^*}\}, \theta)$ \;
$x^{\dagger} = \{x^{\dagger}, \hat{x}^{\dagger^*}\}$ \;
}
\caption{OAT knot selection algorithm. Convergence in the repeat loop is declared when the change in the objective function, the log-marginal likelihood, falls below a threshold. Set 
initial number of knots ($K_I$).}
\label{a:oat}
\end{algorithm}

The precise details of Algorithm \ref{a:oat} will vary depending on the exact model and data set considered. In the case of Gaussian data, the calculation of the log-marginal likelihood is exact. In the case of non-Gaussian data, the Laplace approximation to the marginal likelihood is used.  Initialization of $x^\dagger$
may heavily depend on the task and data under consideration. For example, if $\mathcal{X}$ is one or two dimensional and the 
distribution of the inputs are evenly distributed throughout $\mathcal{X}$, initializing on a grid may work well. In higher dimensions or with inputs that are
non-uniformly distributed, randomly sampling data points may be reasonable. One could also choose the initial set of knots to be the cluster centers 
from a clustering algorithm such as k-means. If the task is classification, one could cluster data separately from each class, which may work better than clustering the combined data. Finally, because the number of initial knots should be small, one could consider simultaneously optimizing the knots
or knots and covariance parameters jointly.  

Several algorithmic choices still remain, such as what algorithm to use to optimize $p(y|x, \{x^{\dagger}, x^{\dagger^*}\}, \theta)$ w.r.t. $(x^{\dagger^*}, \theta)$ and how to choose the knot proposal function $J$, 
which we describe in the next section. Our experience in optimization of 
Gaussian process kernel hyperparameters models is that algorithms like BFGS can 
be fast but numerical instabilities frequently occur. 
In this paper, we use Adadelta \citep{zeiler2012} to do the optimization of 
$p(y|x, \{x^{\dagger}, x^{\dagger^*}\}, \theta)$. While perhaps more commonly seen in 
the deep learning literature, we found Adadelta to be faster and easier to use 
than simple gradient ascent. It also allows us to avoid numerical instabilities due to large step sizes. 

\subsection{Knot Proposal Function}\label{s:proposal_fun}
The choice of $J$ is key to the success of the OAT selection 
method. 
Finding a global optimum over all knots and kernel hyperparameters 
simultaneously is intractable. 
Thus, our modified goal is to find the best possible model over choices of an 
additional knot and kernel hyperparameters given all previously selected knots. 
If knots are restricted to be a subset of $x$, then
one possible way of doing this is to test $x^{\dagger^*} = x_i$ for all 
observed input locations $x_i$ and to select the one which maximizes the 
marginal likelihood. 
Unfortunately, this results in a knot selection and optimization algorithm 
scaling $\mathcal{O}(N^2)$ in computation time. 
It is for this reason that \cite{cao2013} choose to approximate the objective 
function using only a small number of observed data locations
when exploring new possible knot values. 
In order to retain $\mathcal{O}(N)$ scaling, 
we are forced to find a way to cheaply identify candidate knot 
locations which are likely to be good. 
We choose to do this by repurposing the tools of Bayesian optimization (BO)
based on Gaussian processes 
to intelligently explore a small number of highly informative values for a new knot.

BO is a method of finding a global optimum for functions that are expensive to 
evaluate and for which gradient information is usually inaccessible. 
It works by first assuming that the objective function (response surface) of interest can be modeled statistically (usually as a Gaussian process). Locations at which to evaluate the objective function are found sequentially maximizing an 
\emph{acquisition function} at each step. \cite{shahriari2016} provide a review of these methods.
Examples of acquisition functions are probability of improvement or expected improvement \citep[see][for more examples of acquisition functions]{jones2001, shahriari2016}. 
To avoid the confusion due to the fact that we will optimize a 
GP with a GP, 
we will always refer to the GP used to model the log-marginal likelihood as a 
\emph{meta GP}.



Our knot proposal algorithm is described in 
Algorithm \ref{a:oat_proposal}. 
We first define $T_{max}$ to be the maximum number of log-marginal likelihood evaluations that the proposal is allowed to make. 
Then, we randomly select a set of $T_{min}$ potential knots at which we 
evaluate the log-marginal likelihood. 
We then model the log-marginal likelihood values as a function of a new knot 
location with our meta GP, and we optimize the meta GP parameters. 
For each of $T_{max} - T_{min}$ additional time steps, 
we evaluate the log-marginal likelihood at a new new knot which maximizes the 
specified acquisition function,
we use expected improvement \citep{jones1998}, over all data locations. 
We will use notation similar to that in \cite{shahriari2016} 
and denote the acquisition function as $\alpha(z;\cdot, \cdot)$ 
where $z$ denotes the single knot value/vector over which we are trying to optimize the log-marginal likelihood. We explain the other arguments after introducing some additional 
notation. Let $W(z)$ denote the random variable corresponding to the meta GP at input location $z$. Let $w_{1:t-1}$ be the vector of log-marginal likelihood values at the candidates for the knot proposal which have thus far been explored at time $t$. Define $w^{+} = max(w_{1:t-1})$ to be the maximum value of these log-marginal likelihood values. Denote by $\Phi(\cdot)$ the standard normal cumulative distribution function and by $\phi(\cdot)$ the corresponding density function. Expected improvement at step $t$ is given by the expression 

\[
\begin{multlined}
\alpha \left(z;w_{1:t-1}, \left\{x_1^\dagger,\ldots,x_{t-1}^\dagger\right\} \right) = (E\left[W(z)|w_{1:t-1}\right] - w^{+})\Phi\left(\frac{E\left[W(z)|w_{1:t-1}\right] - w^{+}}{\sqrt{ V\left[W(z)|w_{1:t-1} \right] } } \right) + \\
\sqrt{ V\left[W(z)|w_{1:t-1}\right] }\phi \left(\frac{E\left[W(z)|w_{1:t-1}\right] - w^{+}}{\sqrt{ V\left[W(z)|w_{1:t-1}\right] } } \right). 
\end{multlined}
\]

\noindent Note that we have suppressed the dependence of the acquisition function
on the meta GP covariance parameters for notational compactness. After each of the $T_{max} - T_{min}$ iterations, we update the meta GP covariance parameters. At the end of the specified number of iterations, the knot proposal is the knot which resulted in the largest log-marginal likelihood value. In the remainder, we refer to the combination of 
Algorithms \ref{a:oat} and \ref{a:oat_proposal} for knot selection as the OAT-BO
algorithm.

\begin{algorithm}[H]

\SetAlgoLined
set the mean of the meta GP equal to 
$\log p\left(y\left|x,\left\{x^{\dagger},\cdot\right\}, \hat{\theta}\right.\right)$ \;
sample $x_{1 + k}^\dagger, ..., x^\dagger_{T_{min} + k}$ without replacement from $x$ \;

augment known marginal likelihood values 
$w_j = \log p\left(y\left|x,\left\{x^{\dagger},x^{\dagger}_j\right\}, \hat{\theta}\right.\right)$ for $j=1,\ldots,k$ 
with evaluations of the marginal likelihood at the new knots, 
that is $w_{k+j} = \log p\left(y\left|x, \left\{x^{\dagger},x^\dagger_j\right\}, \hat{\theta}\right.\right)$ for $j = 1,...,T_{min}$ \;
\For{$t = T_{min}+1 + k, ..., T_{max} + k$}{
update covariance parameters in meta GP \;
$x^*_t = \argmax_{z \in x \setminus \{x^{\dagger}_l\}_{l = 1}^{t - 1} } \alpha\left(z;w_{1:t - 1},\left\{x_1^\dagger,\ldots,x_{t-1}^\dagger\right\}\right)$ \;
$w_t = \log p\left(y\left|x, \left\{x^{\dagger}, x^*_t \right\}, \hat{\theta}\right.\right)$ \;
}
return $x_j^*$ such that $j=\mbox{argmax}_t w_t$
\caption{Knot proposal algorithm.
Set the minimum ($T_{min}$) and maximum ($T_{max}$) number of marginal 
likelihood evaluations.
}
\label{a:oat_proposal}
\end{algorithm}

A critical component of the proposal function involves taking advantage of the fact that, 
for a new proposed knot $x^{\dagger^*}$ arbitrarily close to an existing knot, 
$p(y|x, \left\{ x^{\dagger}, x^{\dagger^*} \right\}, \theta) \approx p(y|x, x^{\dagger}, \theta)$  
(assuming that any GP noise variance is small). 
That is, when we are modeling the log-marginal likelihood as a function of a new knot, 
we can prespecify the value of the meta GP at the current $k$ knots. 
Furthermore, this value, namely $\log p(y|x, x^{\dagger}, \theta)$, becomes the obvious choice of mean value for the GP because we know that the log-marginal likelihood can take this value when there is no noise variance. 
Embedding this knowledge of the log-marginal likelihood in the meta GP allows it to 
explore parts of the input space efficiently. 

Additionally, if the meta GP has no noise variance, then at time $t$ the expected improvement evaluated at a location $z \in \{x^\dagger_1, ..., x^\dagger_{t - 1} \}$ is zero. In this case, the optimization in Step 6 of Algorithm \ref{a:oat_proposal} is just $x^\dagger_t = \argmax_{z \in x } \alpha\left(z;w,\left\{x_1^\dagger,\ldots,x_{t-1}^\dagger\right\}\right)$. If, however, there is noise variance, then the expected improvement is not zero, and we restrict ourselves to search locations that are yet unexplored.

We finally remark that, if the proposed knot increases the objective function, then the objective function is guaranteed never to decrease in the overall OAT algorithm as long as we do not run into numerical problems during the subsequent gradient based optimizations. Thus, we are given a natural convergence criteria of the change in log-marginal likelihood falling below a specified threshold.


\subsection{OAT Example: Gaussian Data}\label{s:oat_example}

\begin{figure}[htbp!]
\centering
 \includegraphics[width = 1\textwidth, height = 0.3\textheight]{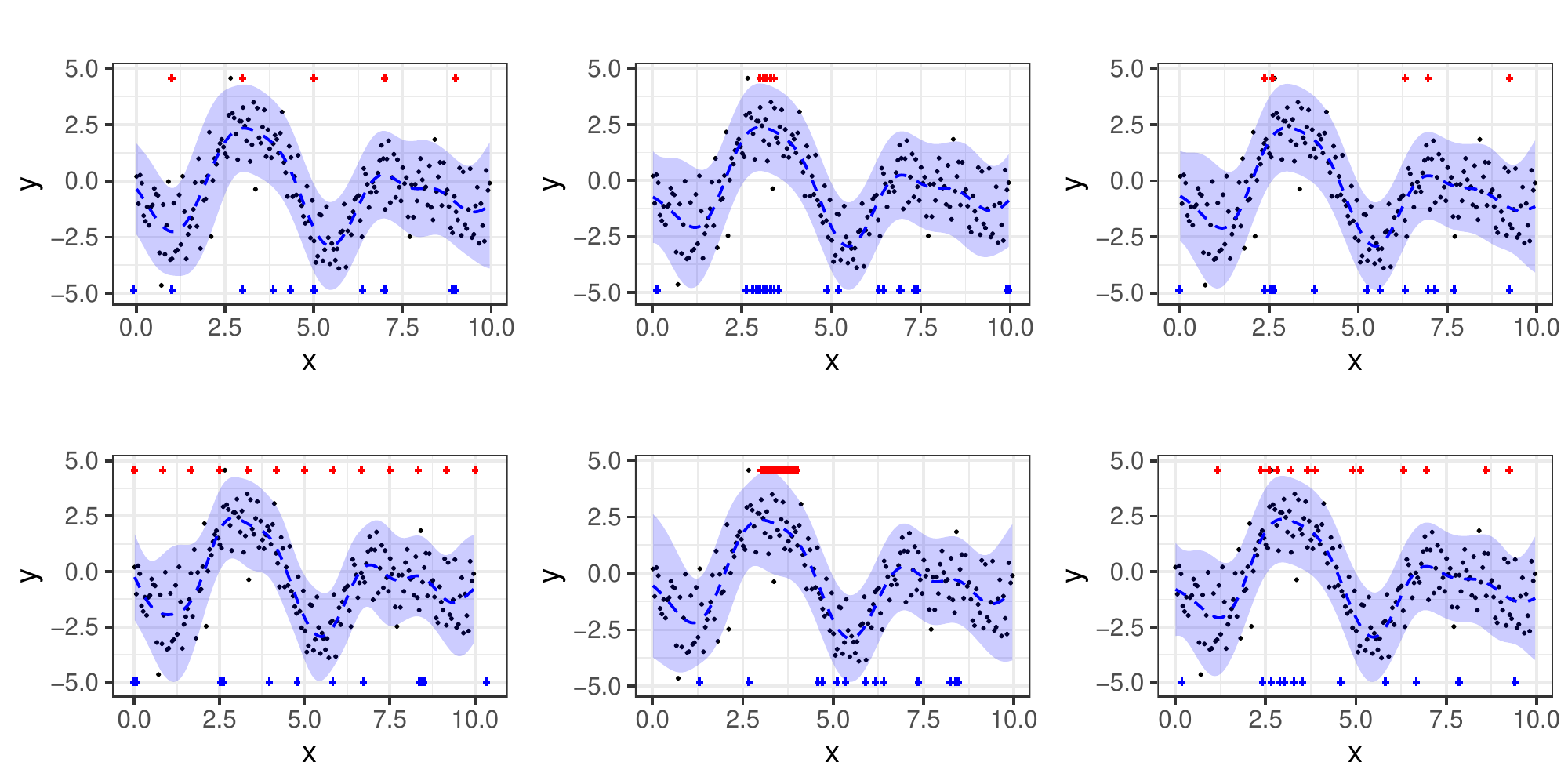}
 \caption{Simulated Gaussian data with posterior predictive means and 95\% prediction intervals for GPs optimized either using the OAT-BO algorithm or where covariance parameters are optimized simultaneously. Three different sets of starting values are used for each optimization. We try a uniform knot initialization, random initialization, and an adversarial initialization where all knots are set close together near the point $x = 3$. The top row of plots corresponds to the OAT knot selection and the bottom row of plots corresponds to simultaneously optimizing all of the knots and covariance parameters. Columns correspond to knot initialization where the leftmost is uniform, middle is adversarial, and right is random. Red crosses on the top of each plot show the initial knot selection, and the blue crosses on the bottom of the plots show the optimized knots.}
 \label{fig:oat_vs_simul_gaussian_example}
\end{figure}

We now consider an illustrative example in the case of Gaussian data. The Gaussian data case is the simplest situation where optimizing knots simultaneously with covariance parameters has already been shown to work well in some situations. We consider 200 simulated observations from a GP with a one dimensional input. Figure \ref{fig:oat_vs_simul_gaussian_example} shows six plots of data with predictions, $E\left[ \left.\tilde{Y}\right|Y \right]$, and $95\%$ prediction intervals, $E\left[ \left.\tilde{Y}\right|Y \right] \pm 1.96 \sqrt{V\left[\left.\tilde{Y}\right|Y\right]}$. Here $\tilde{Y}$ is a new observation at a previously unseen data location, say $\tilde{x}$. The top row corresponds to knots and covariance parameters selected with the OAT algorithm and the bottom corresponds to knots and covariance parameters selected through simultaneous optimization of the log-marginal likelihood. Depending on the initialization, the OAT algorithm selects different numbers of knots. To make comparisons fair, we use the same number of knots selected by the OAT algorithm when optimizing them with the simultaneous method. Red crosses at the top of each plot show the inital set of knots, and the blue crosses at the bottom of each plot show the knots selected at the end of the optimization. The left column corresponds to a uniform knot initialization. The middle column corresponds to an adversarial knot initialization where all knots are set close together near $x = 3$, and the rightmost column corresponds to a random initialization of knots. All covariance parameters are initialized at the true covariance parameter values. For the OAT algorithm, we set $T_{min} = 10$ and $T_{max} = K_{max} = 30$.

In every plot, we see that knots are selected so that they span the input space. Predictions and uncertainties seem largely useful and similar in those cases. In every knot initialization settings, both the OAT selection algorithm and simultaneous optimization result in at least one pair of knots that are close enough together to be redundant. In the OAT case, it is possible that this is still the result of adverse local maxima. 

Table \ref{t:gaussian_example_results} in the supplementary information provides
 some numerical results corresponding to the models in Figure \ref{fig:oat_vs_simul_gaussian_example}.
These include runtime for training, total number of gradient ascent steps,
and RMSEs calculated by taking the differences between the true latent function and the posterior mean. We comment 
that the OAT-BO algorithm always runs more quickly than simultaneous optimization,
and in the case of the adversarial initialization, it is about seven times faster. The runtime difference can be largely explained by the significant reduction in cost of evaluating the gradient. Assuming a one dimensional input space, the derivative with respect to a given knot involves matrix multiplications costing $\mathcal{O}(NK_{max}^2)$. Thus, $K_{max}$ of such matrix multiplications costs $\mathcal{O}(NK_{max}^3)$. For the OAT algorithm, this means that each gradient evaluation is $K_{max}$ times cheaper, implying that OAT can take approximately $K_{max}$ times as many gradient ascent steps without being overall any slower, assuming the knot proposal cost is negligible. The cost of proposing a new knot chiefly involves making predictions and evaluating marginal predictive variances costing $\mathcal{O}(N(T_{max} + K_{max})^2)$ plus some number of matrix inversions coming from the meta GP parameter updates costing $\mathcal{O}((T_{max} + K_{max})^3)$. Thus, the total cost of the proposal function throughout the entire optimization algorithm is roughly bounded above by $T_{max}\mathcal{O}(N(T_{max} + K_{max})^2 + (T_{max} + K_{max})^3)$. Our recommendation for most problems is to make $T_{max} \approx K_{max}$. With $T_{max} = K_{max}$, $T_{max}\mathcal{O}(N(T_{max} + K_{max})^2 + (T_{max} + K_{max})^3) = \mathcal{O}(4NK_{max}^3 + 32K_{max}^4)$. The cost of generating a knot proposal is roughly equivalent to computing $4K_{max}$ knot derivatives plus inversion costs $32\mathcal{O}(K_{max}^4)$. Thus, each new proposal costs fewer than four full gradient evaluations (in the simultaneous optimization case) plus some meta GP matrix inversion costs. Empirically, we observe that the computation time is dominated by gradient evaluations and not the knot proposal costs.

\section{Experiments}\label{s:experiments}
We now consider experiments on three benchmark data sets where the likelihood is Gaussian, Bernoulli, and Poisson. For all experiments, covariance parameters are learned either alongside knots in the case of sparse models, or by themselves in the case of the full GP, using Adadelta \citep{zeiler2012}. We restrict ourselves to data sets small enough that they can be fit with a full GP, but large enough where the computational savings attributed to sparse approximations are noticeable.

We are primarily interested in marginal, as opposed to joint, predictive distributions,
 so we use slightly modified versions of canonical performance metrics. The 
two main metrics we consider are common to all of our experiments. The first metric
is the median negative log-probability (MNLP), which is calculated as 

\[ 
MNLP = \text{median}_{i \in 1, ..., N_{test}} \{ -\log p(\tilde{y}_i|x^{\dagger}, \hat{\theta}, y) \}.
\]

\noindent 
Lower MNLP values correspond to more accurate marginal predictive densities. The second metric 
that we use is the average univariate Kullback-Leibler divergence (AUKL) between the predictive 
density from the full GP versus that of each sparse model. We calculate this as 

\[ 
AUKL = \frac{1}{N_{test}} \sum_{i = 1}^{N_{test}}\int p_{full}(f(\tilde{x}_i)|\hat{\theta}, y) \log \frac{p_{full}(f(\tilde{x}_i)|\hat{\theta}, y)}{p_{sparse}(f(\tilde{x}_i)|x^{\dagger}, \hat{\theta}, y)} df(\tilde{x}_i).
\]

\noindent Values of AUKL near zero indicate that the univariate predictive distributions provided by the 
sparse GP models are close to those of the full GP. 






As a benchmark for Algorithm \ref{a:oat_proposal}, on each data set we fit FIC models using the OAT algorithm with a proposal function which is the best of a random subset of the data locations. We will abbreviate this proposal function as RS for \textit{random subset}. That is, to propose a knot, we sample $x^{*} \subset x$ with $\lvert x^{*} \rvert = T_{max}$. Then, the RS proposal is $x^{\dagger^*} = \argmax_{i \in 1,...,T_{max}} p(y|x, \{x^{\dagger} , x^{*}_i\}, \theta)$. 

Starting values for covariance parameters were chosen separately for each experiment but kept consistent for each GP model fit. In all experiments, a noise variance parameter with a positive but small, lower bound was estimated. 

\subsection{Boston Housing Data}\label{s:boston}
We first consider the Boston housing data set and use ``\% lower status of the population'', ``average number of rooms per dwelling'' and ``pupil-teacher ratio by town'' to predict the median value of owner occupied homes. We removed observations where the median value was less than \$50,000. This resulted in 490 observations which we randomly split $\approx$ 80/20 into training and testing data. We fit a full GP model, an FIC model using the OAT algorithm with a proposal function given by Algorithm \ref{a:oat_proposal} (OAT-BO), two FIC models using the RS knot proposal function (OAT-RS) with different values of $T_{max}$, and two FIC models where the knots and covariance parameters were simultaneously optimized. Of the two FIC models fit using simultaneous optimization, one is fit using our maximum number of knots $K_{max} = 50$ while the other is fit using the number of knots selected by OAT-BO. Knot locations for all sparse models were initialized using k-means clustering. For models fit using OAT knot selection, the intial number of knots was set to five.

Table \ref{t:boston_results} provides runtimes in seconds as well as accuracy in the form of a standardized root mean squared error, as well as the median negative log-probability. The SRMSE calculates the root mean squared error between the GP predictions and the test data normalized by the sample standard deviation on the test set. That is,

\[ SRMSE = \sigma_{\tilde{y}}^{-1} \sqrt{ \frac{1}{N_{test}} \sum_{i = 1}^{N_{test}} (E\left[f(\tilde{x}_i)|Y\right] - \tilde{y}_i )^2}, \]

\noindent 
where $\sigma_{\tilde{y}}^2 = \frac{1}{N_{test} - 1}\sum_{i = 1}^{N_{test}}(\tilde{y}_i - \ubar{\tilde{y}})^2$, $\ubar{\tilde{y}} = \frac{1}{N_{test}} \sum_{i = 1}^{N_{test}}\tilde{y}_i$, and $\tilde{y}$ is the vector of test set target values.

The most striking result here is how long the simultaneous optimizations took to complete. We limited each optimization to 1000 iterations, at which point neither simultaneously optimized model had met our convergence criteria (though they did appear to be close). Both the full GP and the sparse model fit with OAT met the relevant convergence criteria. The sparse models fit using OAT all had runtimes that were somewhat similar to the full GP, with the OAT-BO model taking about 100 seconds longer. Conversely, the simultaneous models both took longer to fit than the full GP. When $K = 50$, optimization took roughly 65 times longer to fit than the full GP and about 10 times longer when $K = 13$. 

One might expect, despite this seeming convergence issue, that the accuracy of 
predictions of the model with 50 knots would surpass that of the sparser models.
However, we see better predictive performance with all of the sparser models. 
All models do well in terms of accuracy in predicting the actual test target 
values. The worst model (with 50 knots) has a SRMSE value of $0.378$ compared to
$0.359$ for the full GP. The sparse model that reproduces predictions from the 
full GP best is the OAT-RS model with $T_{max} = 25$ which has a AUKL value of
$0.039$, though both other OAT models are not far behind. Both simultaneously 
optimized models tend to do worse in terms of reproducing the 
predictive distribution of the full GP, and the 50 knot model was the worst in this regard.

Interestingly, the MNLP values seem to indicate that the best models are the two
sparse models where knots are simultaneously optimized. This, however, largely seems
to be due to the differences in predictive uncertainty compared to 
the full GP as indicated by the AUKL metric. This appears to be an undesirable 
idiosyncrasy as one may have uncertainties that tend to be poorly calibrated in 
certain regions of the input space.

\begin{table}
\centering
\begin{tabular}{|l|r|r|l|r|r|r|}
\hline
Method & Runtime & K & Tmax & SRMSE & MNLP & AUKL\\
\hline
Full & 394 & -- & -- & 0.359 & 2.500 & 0.000\\
\hline
OAT-BO & 545 & 13 & 25 & 0.366 & 2.466 & 0.045\\
\hline
OAT-RS & 356 & 12 & 25 & 0.366 & 2.464 & 0.039\\
\hline
OAT-RS & 339 & 15 & 50 & 0.364 & 2.469 & 0.047\\
\hline
Simult. & 25831 & 50 & -- & 0.378 & 2.291 & 0.356\\
\hline
Simult. & 3945 & 13 & -- & 0.356 & 2.313 & 0.242\\
\hline
\end{tabular}
\caption{Results on the \textit{Boston Housing} data. Runtimes (in seconds) needed to train each model are shown. The SRMSE calculates the root mean squared error between the predictions from each GP and the target values and is divided by the standard deviation of the actual test set target values. MNLP calculates the median negative log-probability of each test observation. AUKL calculates the average univariate KL divergence between the predictive distributions of the latent function coming from the full versus the sparse GP models. Number of knots, $K$, and the number of objective function evaluations per proposal, $T_{max}$, is also shown for each model.
}
\label{t:boston_results}
\end{table}

\subsection{Banana Data}\label{s:banana}
To showcase the OAT algorithm's performance on a binary classification task, we fit a full GP model as well as four sparse GP models to the \textit{Banana} data set. In order to enable fitting of the full GP model, we used a random subset of only 531 cases, or about 10\% of the data, for training and the remaining 4769 cases were used as the test set. In both sparse models, knots were initialized using k-means clustering. For models fit using OAT knot selection, the initial number of knots was set to five.

\begin{figure}[htbp!]
\centering
 \includegraphics[width = 1\textwidth, height = 0.5\textheight]{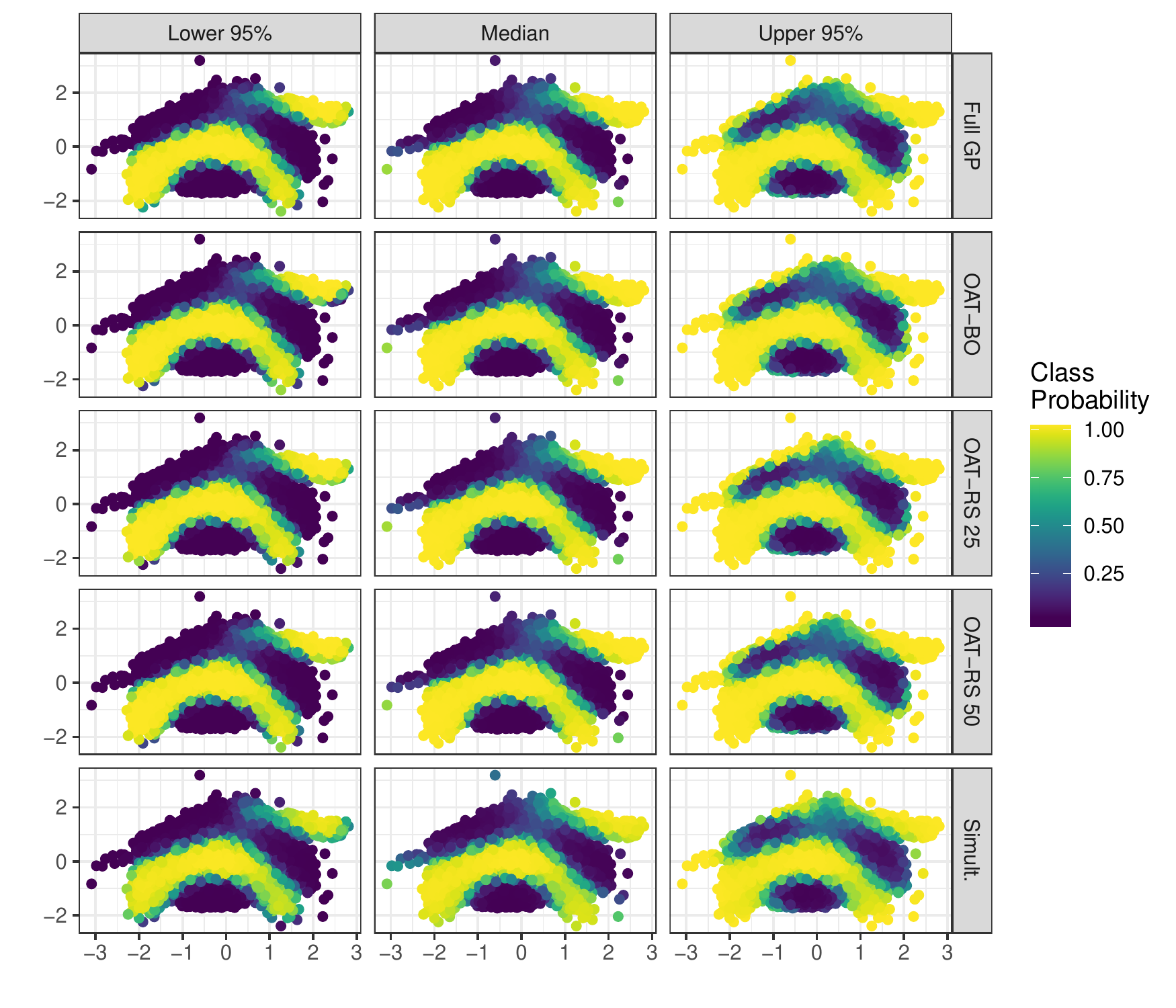}
 \caption{Estimated posterior class probabilities with 95\% credible intervals on the Banana data test set for the full GP fit, sparse GP fit with the OAT algorithm, and sparse GP fit where all knots and covariance parameters are optimized simultaneously.}
 \label{fig:banana_test_set_plot}
\end{figure}

Figure \ref{fig:banana_test_set_plot} shows estimates of probabilities for class one membership along with 95\% credible intervals for the full GP, sparse GP fit with the OAT knot selection algorithm using both Algorithm \ref{a:oat_proposal} and RS as proposal functions, and sparse GP fit by simultaneously optimizing knots and covariance parameters with $K = 50$. All sparse models choose to include all 50 knots. The estimated probabilities between the full GP and the sparse models are visually nearly indistinguishable.


\begin{figure}[htbp!]
\centering
 \includegraphics[width = 0.5\textwidth, height = 0.15\textheight]{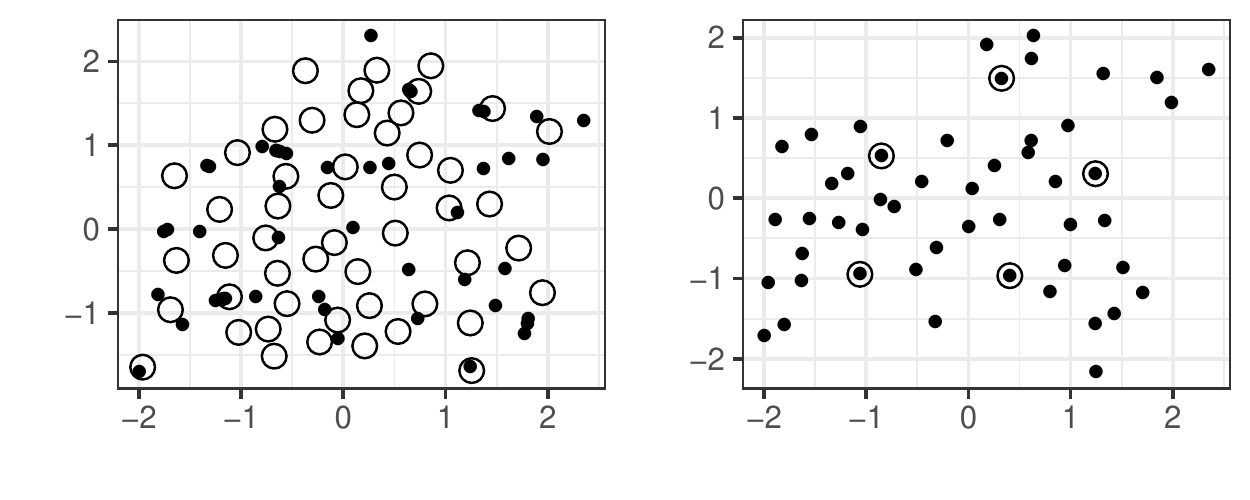}
 \caption{Locations of initialized and estimated knots for the simultaneously optimized model with 50 knots (left) and for the OAT-BO model (right). Open circles are initial knots and solid points are estimated knots.}
 \label{fig:banana_knot_alloc}
\end{figure}

Table \ref{t:banana_results} shows runtimes in seconds as well as measures of predictive performance for the five models. As suggested from Figure \ref{fig:banana_test_set_plot}, all sparse models fit using OAT knot selection come close to the predictive performance of the full GP. The worst model is the one fit by simultaneously optimizing the knots, which has a value of MNLP that is
nearly twice that of the other models. The AUKL values for the OAT models are all
similarly small, while the value for the simultaneously optimized model is roughly 50-100 times larger than any given OAT model. This can be explained by the poor selection of knots as shown in the left panel of Figure \ref{fig:banana_knot_alloc}, where there is more clumping of knots. Thus, estimates of uncertainty are more likely to be different than for the more even knot distributions of the OAT models. Fitting sparse models is less time consuming than fitting the full GP, but both the OAT-BO and OAT-RS with 
$T_{max} = 25$ models fit in approximately half of the time that it takes the non-OAT model.

\begin{table}
\centering
\begin{tabular}{|l|r|l|r|r|r|}
\hline
Method & Runtime & Tmax & K & MNLP & AUKL\\
\hline
Full & 26795 & -- & -- & 0.038 & 0.000\\
\hline
OAT-BO & 3150 & 25 & 50 & 0.037 & 0.061\\
\hline
OAT-RS & 2954 & 25 & 50 & 0.038 & 0.051\\
\hline
OAT-RS & 3471 & 50 & 50 & 0.038 & 0.039\\
\hline
Simult. & 6219 & -- & 50 & 0.069 & 3.265\\
\hline
\end{tabular}
\caption{Results on the \textit{Banana} data. Runtimes (in seconds) needed to train each model are presented. MNLP calculates the median negative log-probability of each test observation. AUKL calculates the average univariate KL divergence between the predictive distributions of the latent function coming from the full versus the sparse GP models. Number of knots, $K$, and the number of objective function evaluations per proposal, $T_{max}$, is also shown for each model.}
\label{t:banana_results}
\end{table}

\subsection{Lansing Woods Hickory Data}\label{s:lansing}
In this example, we model the counts of hickory trees in Lansing Woods in Michigan on an even $30 \times 30$ grid. These data are available from the {\tt spatstat.data} R package \citep{spatstat.data}. 

We treat this as a smoothing, rather than a prediction problem. We fit a full GP model as well as five sparse models, and our performance measures compare the fit of the full GP to that of sparse models on the data used to fit the models. We use the formulas for prediction derived in Section \ref{s:nongaussian_data}
in order to get smoothed estimates of the intensity at the \emph{observed} data locations. 
Formulas were derived for \emph{unobserved} data locations, but these are the 
appropriate formulas to use here as well. 

We initialized knots uniformly in the cases where knots were simultaneously optimized, and they were initialized to a random subset of 10 data locations when using OAT knot selection.

\begin{figure}[htbp!]
\centering
 \includegraphics[width = 1\textwidth, height = 0.5\textheight]{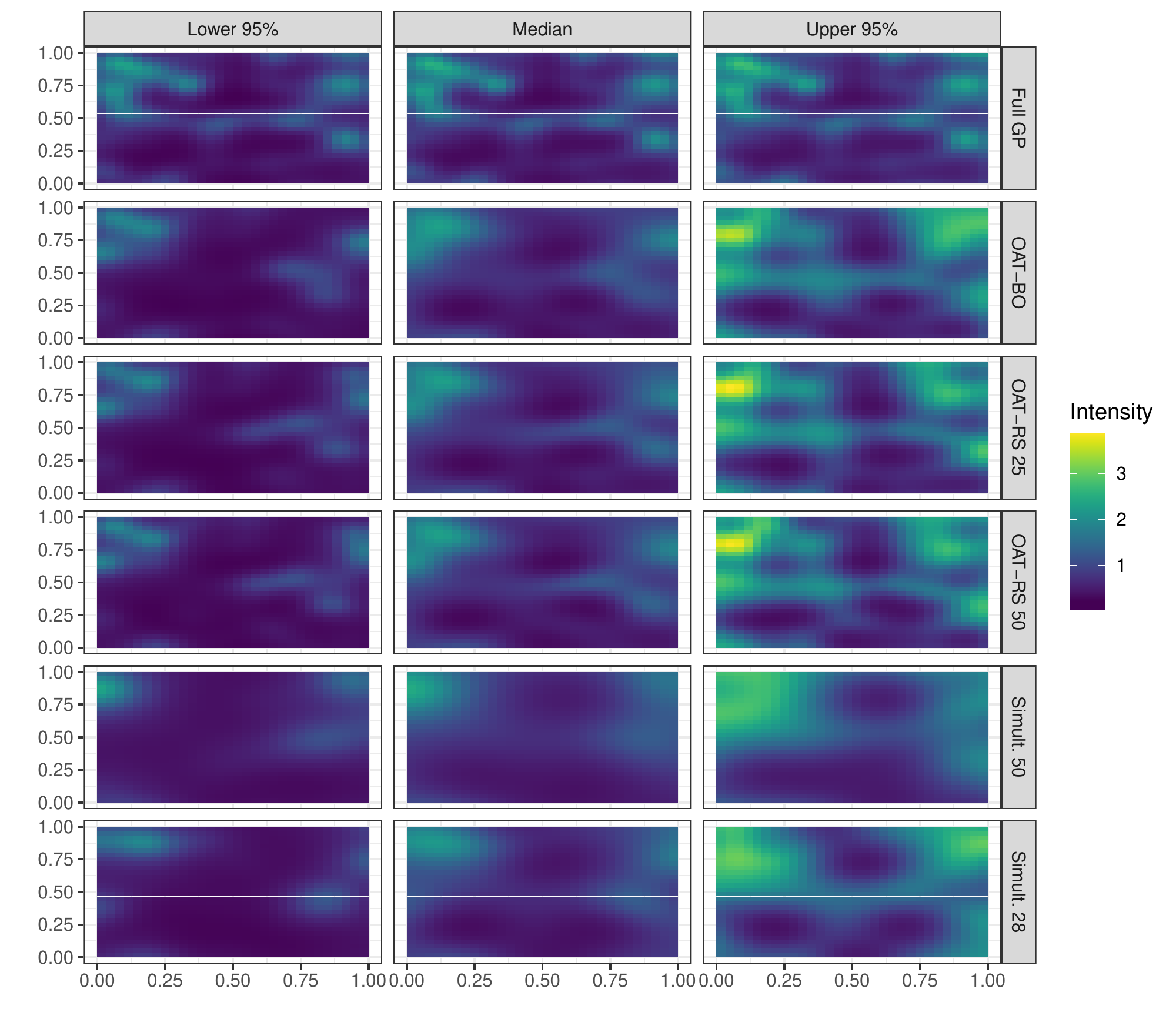}
 \caption{Estimated posterior class probabilities with 95\% credible intervals on the Hickory data test set for the full GP fit, sparse GP fit with the OAT algorithm, and sparse GP fit where all knots and covariance parameters are optimized simultaneously.}
 \label{fig:hickory_plot}
\end{figure}

\begin{figure}[htbp!]
\centering
 \includegraphics[width = 0.5\textwidth, height = 0.15\textheight]{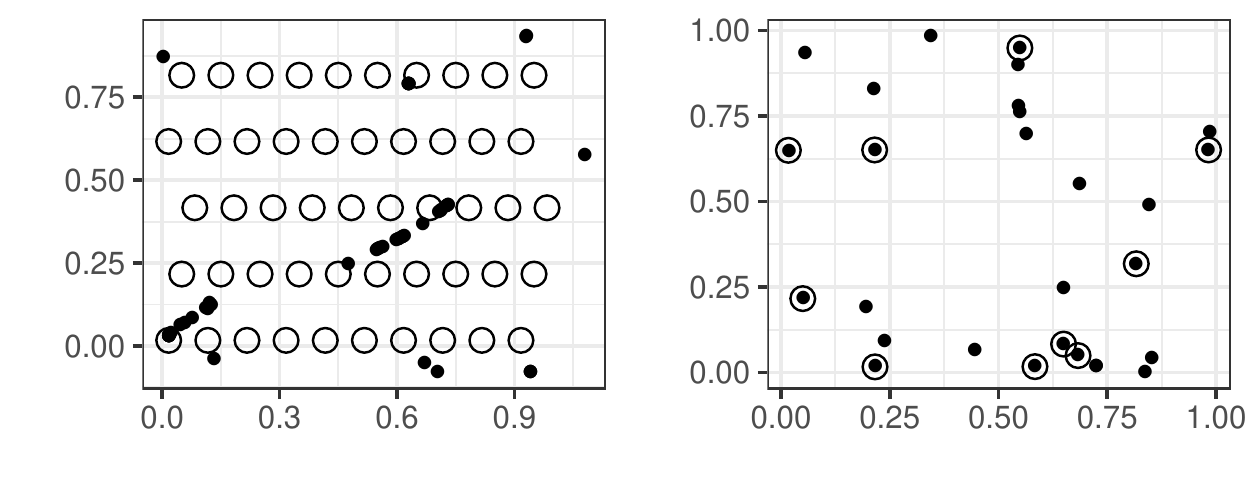}
 \caption{Locations of initialized and estimated knots for the simultaneously optimized model with 50 knots (left) and for the OAT-BO model (right). Open circles are initial knots and solid points are estimated knots.}
 \label{fig:hickory_knot_alloc}
\end{figure}

Figure \ref{fig:hickory_plot} shows the estimated (posterior median) intensities as well as 95\% credible intervals. All of the sparse models over-smooth the data, yet once again, the worst model is the one with the largest number of knots. Figure \ref{fig:hickory_knot_alloc}, again, shows why the performance of the simultaneously optimized model is so poor. The knot allocation here are reminiscent of that shown for the Banana data in Figure \ref{fig:banana_knot_alloc}, but the situation is worse here. Most of the 50 knots are pushed near to each other in the center of the domain. Selecting knots using the OAT procedure avoids this issue entirely.

As expected from the visual analysis of the estimated intensities, the AUKL values for the sparse models are all smaller than those of the simultaneously optimized models, though both OAT-RS models appear to do slightly better than the OAT-BO model. In the worst case, the OAT-RS model with $T_{max} = 25$ has an AUKL value almost half that of the 50 knot model. 
MNLP values are all better for the OAT models than for the simultaneously optimized
model with 50 knots, and the OAT-BO and OAT-RS with $T_{max} = 50$ had the closest
MNLP values to the full GP.

All OAT models fit close to five times faster than the full model, with the simultaneously optimized models both taking longer than the full GP. The 50 knot model takes over three times as long to fit as the full GP while the 28 knot, simultaneouslty optimized model takes just under three times longer.


\begin{table}
\centering
\begin{tabular}{|l|r|r|l|r|r|}
\hline
Type & Runtime & K & Tmax & MNLP & AUKL\\
\hline
Full & 5892 & -- & -- & 1.040 & 0.000\\
\hline
OAT-BO & 846 & 28 & 25 & 1.058 & 0.321\\
\hline
OAT-RS & 1083 & 33 & 25 & 1.066 & 0.276\\
\hline
OAT-RS & 989 & 30 & 50 & 1.055 & 0.279\\
\hline
Simult. & 18456 & 50 & -- & 1.068 & 0.597\\
\hline
Simult. & 15046 & 28 & -- & 1.060 & 0.426\\
\hline
\end{tabular}
\caption{Results on the \textit{Hickory} data including runtimes (in seconds) needed to train each model. MNLP calculates the median negative log-probability of each test observation. AUKL calculates the average univariate KL divergence between the predictive distributions of the latent function coming from the full versus the sparse GP models. Number of knots, $K$, and the number of objective function evaluations per proposal, $T_{max}$, is also shown for each model.}
\label{t:hickory_results}
\end{table}

\section{Discussion} \label{s:discussion}
We proposed a method for selecting the number and locations of knots in sparse Gaussian process models based on optimizing the marginal likelihood in the case of Gaussian data, or an approximation of it when the data are not Gaussian. Our method allows for sparse approximations to effectively adapt to the complexity of the function being modeled. Experiments on several benchmark data sets yield as good or better performance than simultaneous optimization of knots and covariance parameters. Furthermore, the performance of the resulting models is often comparable to that of a full GP, but with computational savings. 

One of the key reasons why our method outperforms simultaneous optimization of all knots and covariance parameters is that it avoids optimization problems inherent to the likelihood surface. As a function of knots, the marginal likelihood is riddled with suboptimal local maxima, many of which occur when multiple knots are close enough together to be practically redundant. We see this behavior in synthetic as well as real world data, and while our knot selection algorithm is not guaranteed to avoid the problem completely, in practice performance is improved. Furthermore, similar problems ought to be encountered when optimizing other marginal likelihood approximations as a function of knots as in, for example, GP classification with expectation propagation (EP). Thus, we think that our OAT knot selection scheme could result in performance improvements there as well.

\cite{bauer2016} argue that the success of FI(T)C 
implementations where the knots are simultaneously and continuously optimized
is often a result of local, rather than global, optima. OAT knot selection
does not even attempt to find a global optimum, and so using the OAT-BO algorithm
appears to be an effective, practical way to find a local optimum in a principled way:
by finding a sequence of approximately optimal models holding the knots found
previously fixed. Furthermore, using OAT knot selection, we have not encountered
the dramatic underestimation of noise variance observed in \cite{bauer2016}. 




We propose using Bayesian optimization to efficiently search for candidate knot proposals.
Interestingly, using Bayesian optimization to search for candidate knot proposals yielded performance that was often no better than using the RS proposal. This fact underscores the notion that it is often not hard, at least while the approximation is poor, to find knots that improve the marginal likelihood when added to the current model, and this is supported by \cite{bauer2016}. 

However, this also raises the question as to why RS was as competitive as it was. One possibility is that the RS proposals tended to be worse, in terms of increasing the marginal likelihood, but that this may be desirable as the final approximations may use more knots and be closer to the fit
of the full GP. Unfortunately, this would be a nail in the coffin for 
sophisticated greedy knot selection strategies that use the marginal likelihood as the objective function,
as it would be unclear how to propose a new knot. 

Another possibility is that the BO proposal spends too much time exploring suboptimal local maxima and thus explores fewer knots that are as far apart as the RS proposal does. One could increase the number of likelihood evaluations allowed during the knot proposal process at additional computational cost, but a better option, at least in theory, would be to account for the finite set of marginal likelihood queries in the acquisition function. Such an accounting has been studied in \cite{ginsbourger2009}, who show that the appropriately modified acquisition function can result in more exploration throughout the domain. However, the methods proposed in their paper are computationally prohibitive for even very small values of $T_{max}$. \cite{gonzalez2016} developed a approximate method for non-myopic Bayesian optimization. Their algorithm, or a variant of it, could potentially improve upon the proposal function used in this work. A final option would be to forego the BO proposal mechanism altogether and opt for cheap, model-specific approximations to identify candidate knots that are likely to be good.

While the optimization problems inherent to the marginal likelihood may not be as severe for the optimization objective used during variational inference as in \citep{titsias2009}, there are still plenty of suboptimal local maxima \citep{bauer2016}. Even as a function of a single knot, \cite{bauer2016} 
show that there are necessarily multiple optima due to ``spikes" observed when 
a knot is added exactly to a previous knot location. Thus, we feel that it may 
be possible to find improved approximations at lower computational cost
by using OAT knot selection in this setting as well.   

Finally, there are two main weaknesses with the OAT algorithm. OAT seems to do well when length scales are long relative to the size of the domain, but it may terminate early when length scales are short. We saw this in the Lansing woods example where the OAT algorithm avoided optimization issues encountered through simultaneous optimization of the knots, but it also terminated before reaching the maximum number of allowable knots even though the resulting estimate of the intensity was smoother than the full GP. The other weakness is that the OAT algorithm tends to select knots only until the point estimate of the latent function is a good approximation of the full GP. This can lead to OAT terminating even when there are areas of the domain with inflated uncertainties. Both of these problems might be amenable to modifications of the marginal likelihood objective function (such as through an appropriate hyperprior), which we plan on exploring in the future.


\section*{Acknowledgements}
This work was partially funded by the 452 Center for Statistics and Applications
in Forensic Evidence (CSAFE) 453 through Cooperative Agreement \#70NANB15H176 
between NIST 454 and Iowa State University, which includes activities carried out
at 455 Carnegie Mellon University, University of California Irvine, and 456 University of Virginia.

This work was also partially funded by the Iowa State University Presidential Interdisciplinary Research Initiative on C-CHANGE: Science for a Changing Agriculture.



\bibliography{genref_dissertation}

\section{Appendix}

\subsection{Simulated Gaussian Results}
To measure the accuracy of the resulting predictions, we compute root mean squared errors (RMSEs) comparing the posterior predictive means to the simulated GP mean values $f_x$. Specifically,

\[ 
\text{RMSE}(\hat{f}_x,f_x) = \left[\frac{1}{N}\sum_{i = 1}^{N}{\left(\hat{f}(x_i) - f(x_i)\right)^2} \right]^{1/2},
\]

\noindent where 

\[ 
\hat{f}(x_i) = \hat{\Sigma}_{\tilde{x}x^{\dagger}} \hat{\Sigma}_{x^{\dagger}x^{\dagger}}^{-1} \left( \hat{\Sigma}_{x^{\dagger}x} (\hat{\Psi}_{xx} + \hat{\tau}^2I)^{-1}(y - m_x) \right) + m(x_i).
\]

\noindent The expression for $\hat{f}(x_i)$ comes from explicitly using the law of iterated expectation, namely $\hat{f}(x_i) = E\left[f(x_i)|y \right] = E\left[ E \left[ f(x_i) | f_{x^{\dagger}} \right] | y \right]$. We set $m(x_i) = 0$. Table \ref{t:gaussian_example_results} shows the RMSE, runtime, and log-marginal likelihood values for each of the situations shown in Figure \ref{fig:oat_vs_simul_gaussian_example}. 
We also provide the RMSE for predictions from a full GP using the true covariance parameter values. 

\begin{table}
\centering

\begin{tabular}{|l|l|r|r|r|r|r|}
\hline
Method & Initialization & K & RMSE & Runtime & GA Steps & log-Likelihood\\
\hline
Full GP & -- & -- & 0.192 & -- & -- & -311.720\\
\hline
OAT & Uniform & 13 & 0.180 & 50 & 464 & -308.120\\
\hline
OAT & Adversarial & 22 & 0.228 & 96 & 669 & -308.587\\
\hline
OAT & Random & 13 & 0.228 & 50 & 470 & -308.225\\
\hline
Simult. & Uniform & 13 & 0.220 & 140 & 212 & -306.852\\
\hline
Simult. & Adversarial & 22 & 0.196 & 700 & 529 & -308.398\\
\hline
Simult. & Random & 13 & 0.247 & 88 & 140 & -308.071\\
\hline
\end{tabular}
\caption{RMSEs, runtimes (seconds), and optimized log-marginal likelihood values for GPs optimized either using the OAT-BO algorithm or where covariance parameters are optimized simultaneously. Also included is the RMSE from using a full GP fit and using the true covariance parameter values. Three different sets of starting values are used for each optimization. We try a uniform knot initialization, random initialization, and an adversarial initialization where all knots are spaced $0.1$ apart near the point $x = 2$. RMSEs are calculated by summing the squared distances from the actual simulated mean function at the data locations.
}
\label{t:gaussian_example_results}
\end{table}

We see that the range of RMSE values is similar between the simultaneous and OAT models. The best RMSE is the OAT model with uniform initialization, which is $0.18$ compared to $0.192$ for the full GP using the true covariance parameters. The fact that the sparse model has smaller RMSE could be an indication of slight overfitting.
Despite arguably similar accuracy, the OAT algorithm provides a speed advantage, especially in the adversarial initialization when the simultaneous optimization is more than seven times slower. In this situation, the OAT algorithm adapts to the bad initialization by placing more knots than it deems necessary with a better initial set of points. For both the uniform and random initializations, the OAT algorithm selects the exact same number of knots, though the locations are different. We do not expect this to happen in general, but it suggests that OAT knot selection might consistently choose the appropriate degree of sparsity. The performance of the OAT knot selection method in the adversarial initialization is even more promising considering that five knots are fixed at the outset, while the simultaneous method is allowed to optimize all knots. Of course, this difference adds to the speed advantage of OAT as well, but as long as the accuracy does not meaningfully suffer, we feel that it is still a fair point to make.

\subsection{OAT Example: Poisson Data} \label{s:poisson_example}
We now consider the case where the data are Poisson distributed. Specifically, we suppose that for $i = 1,...,399$,

\begin{equation*}
Y_i | f(x_i) \stackrel{ind}{\sim} \text{Poisson}(\lambda_i a_i),
\end{equation*}

\noindent where $a_i = a = x_{2} - x_{1}$ and $\lambda_i = e^{f(x_i)}$. This model corresponds to an approximation of a log-Gaussian Cox process \citep{moller1998} where the intensity is approximated by a constant for each small interval of the input space of length $a$. 

In the non-Gaussian data case, recall that we select knots and covariance parameters by optimizing the Laplace approximation to the log-marginal likelihood. Deriving this approximation also provides a simple Gaussian approximation to the posterior distribution of any latent function values, say $f_{\tilde{x}}$, at unobserved data locations $\tilde{x}$. For simplicity, we use this Gaussian approximation when showing predictions and comparing our OAT knot selection algorithm to the simultaneous optimization of knots and covariance parameters. However, other approximate inference methods, such as MCMC, could be used instead. 

The purpose of this example is to compare results of the OAT algorithm to simultaneous optimization of knots. Additionally, we use this opportunity to visually inspect how optimizing knots by either means differs from a fixed knot and a full GP model. We compare six different models. The first two models are OAT models with two different initializations of three knots and a computational limit of $K_{max} = 20$ knots. We set $T_{min} = 10$ and $T_{max} = 20$. The second two models are sparse with $K = 11$ and $K = 20$ where the knots are optimized simultaneously. The fifth model is a sparse model with $K = 20$ evenly spaced knots, and the sixth model is a full GP. For each model, covariance parameters were optimized and used the same starting values. 

\begin{figure}[htbp!]
\centering
 \includegraphics[width = 1\textwidth, height = 0.4\textheight]{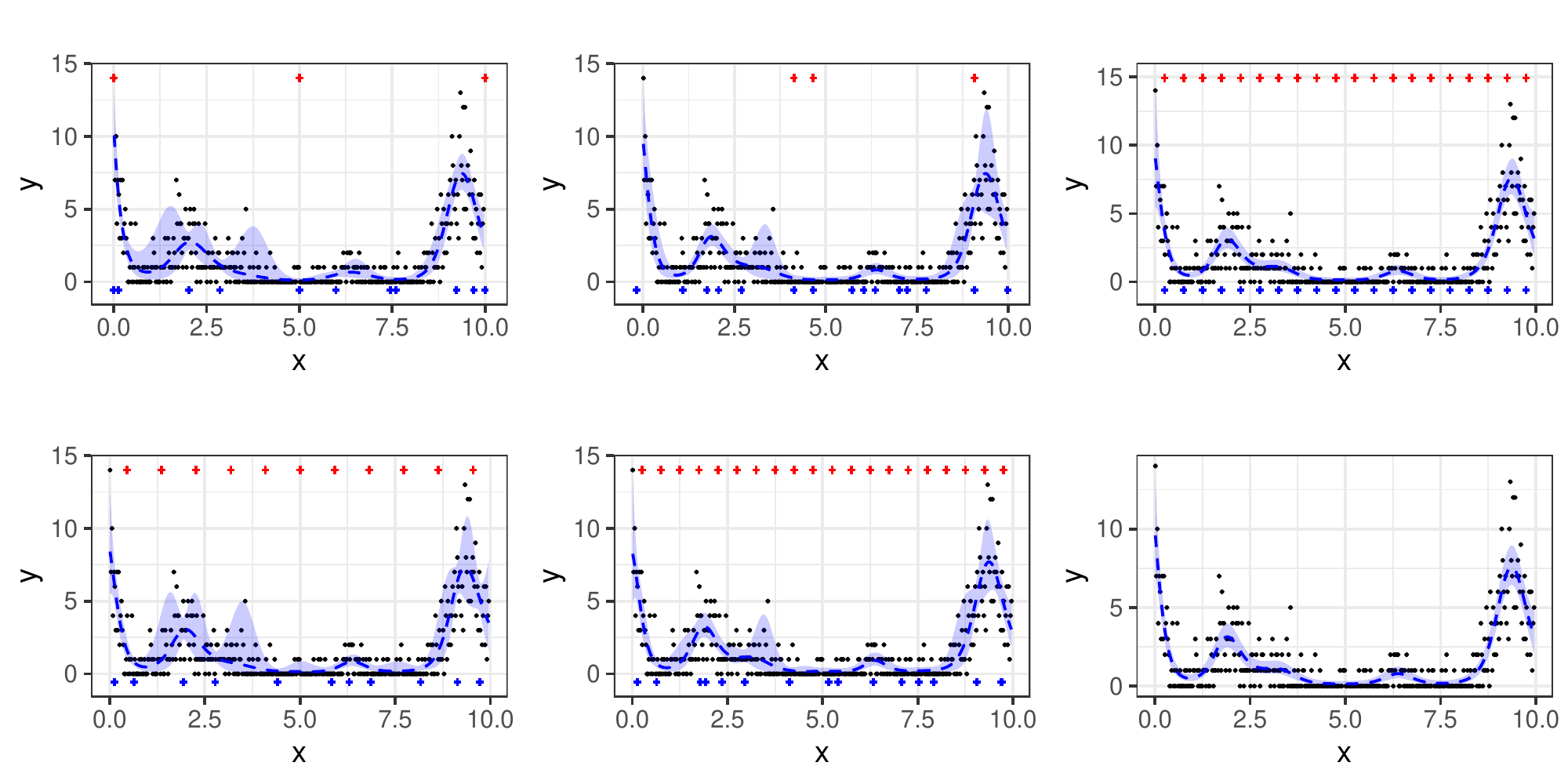}
 \caption{Simulated Poisson data with posterior predictive means and 95\% credible intervals for GPs optimized either using the OAT-BO algorithm or where covariance parameters are optimized simultaneously. We try a uniform knot initialization and random initialization when using the OAT-BO algorithm. Only uniform initializations are used for simultaneous optimization of the covariance parameters and knots (where $K = 11,20$). This is compared to a fit with 20 evenly spaced knots and optimized covariance parameters. RMSEs are calculated by summing the squared distances from the predicted log-intensities to the actual log-intensity function at the data locations. Red crosses on the top of each plot show the initial knot selection, and the blue crosses on the bottom of the plots show the optimized knots.}
 \label{fig:oat_vs_simult_poisson_example}
\end{figure}

Figure \ref{fig:oat_vs_simult_poisson_example} shows the simulated Poisson data along with predictions of the intensity functions ($e^{E\left[ f_{\tilde{x}}|Y \right] a}$) and now asymmetric $95\%$ credible intervals \\
\noindent ($e^{(E\left[ f_{\tilde{x}}|Y \right] \pm 1.96 \sqrt{V\left[f_{\tilde{x}}|Y\right]}) a}$). The top left and top middle plot show results for the OAT algorithm with three knot uniform and random initializations, respectively. With the uniform initialization, the OAT algorithm selected 11 total knots, and with the random initialization, the OAT algorithm selected 19 total knots. The bottom left and middle plots show results from the simultaneous optimization of knots and covariance parameters with 11 and 20 knots, respectively. The top right plot shows a fit with 20 evenly spaced knots which were held fixed while the covariance parameters were optimized. The bottom right plot shows the fit of a full GP with optimized covariance parameters. 

Here OAT tends to place knots evenly throughout the domain. Here we also see that initialization can impact the sparsity of the resulting approximation, as the random knot initialization has seven additional knots than the uniform initialization. Despite this, the point estimates for the intensity are visually nearly identical between both OAT models and also when compared to the results from the other models.

The biggest difference between each model is in the posterior uncertainty of the latent function. All models which optimize knots overestimate uncertainty typically near $x = 2,3,9$. We can expect to occasionally pay this price in these types of models. However, this price is unnecessary in this case, as the interval estimates for the intensity when we fix 20 evenly spaced knots are nearly identical to the full GP.




\begin{table}
\centering
\begin{tabular}{|l|r|l|r|r|r|r|}
\hline
Method & K & Knot Init. & RMSE & Runtime & GA Steps & log-Likelihood\\
\hline
Full GP & -- & NA & 0.198 & 195 & 158 & -511.006\\
\hline
OAT & 11 & uniform & 0.249 & 135 & 950 & -510.715\\
\hline
OAT & 19 & random & 0.188 & 177 & 958 & -508.356\\
\hline
Simultaneous & 11 & uniform & 0.187 & 226 & 194 & -505.931\\
\hline
Simultaneous & 20 & uniform & 0.202 & 197 & 79 & -507.725\\
\hline
Fixed & 20 & fixed uniform & 0.197 & 17 & 166 & -510.909\\
\hline
\end{tabular}
\caption{RMSEs, runtimes (seconds), and optimized objective function values for latent GPs (with Poisson data) optimized either using the OAT-BO algorithm or where covariance parameters are optimized simultaneously alongside knots. Point estimates and $95\%$ credible intervals are generated via a predictive Gaussian approximation to $f_x|y,x^{\dagger},\theta$. We try a uniform knot initialization and random initialization when using the OAT-BO algorithm. Only uniform initializations are used for simultaneous optimization of the covariance parameters and knots (where $K = 11,20$). This is compared to a fit with 20 evenly spaced knots and optimized covariance parameters as well as the fit of a full GP with optimized covariance parameters. RMSEs are calculated by summing the squared distances between the estimated and the actual log-intensity at the data locations.}
\label{t:poisson_example_results}
\end{table}

\end{document}